\documentclass[sn-nature,iicol]{sn-jnl}% Style for submissions to Nature Portfolio journals
\usepackage{graphicx}%
\usepackage{multirow}%
\usepackage{amsmath,amssymb,amsfonts}%
\usepackage{amsthm}%
\usepackage{mathrsfs}%
\usepackage[title]{appendix}%
\usepackage{xcolor}%
\usepackage{textcomp}%
\usepackage{manyfoot}%
\usepackage{booktabs}%
\usepackage{algorithm}%
\usepackage{algorithmicx}%
\usepackage{algpseudocode}%
\usepackage{listings}%
\theoremstyle{thmstyleone}%
\theoremstyle{thmstyletwo}%

\theoremstyle{thmstylethree}%

\begin{document}

% \title[Article Title]{Informational Embodiment: Information Structure for Efficient Codes and Bodies}
% \title[Article Title]{Informational Embodiment: Computational role of information structure in robot sensing, control and body design}

\title[Article Title]{Informational Embodiment: Computational role of information structure in codes and robots}

%\title[Article Title]{The Informational Body: Efficient Codes for Unreliable Robots}


\author*[1]{\fnm{Alexandre} \sur{Pitti}}\email{alexandre.pitti@ensea.fr}

\author[2]{\fnm{Kohei} \sur{Nakajima}}%\email{iauthor@gmail.com}
\equalcont{These authors contributed equally to this work.}

\author[2]{\fnm{Yasuo} \sur{Kuniyoshi}}%\email{iiiauthor@gmail.com}
\equalcont{These authors contributed equally to this work.}

%\affil*[1]{\orgdiv{Department}, \orgname{Organization}, \orgaddress{\street{Street}, \city{City}, \postcode{100190}, \state{State}, \country{Country}}}

%\affil[2]{\orgdiv{Department}, \orgname{Organization}, \orgaddress{\street{Street}, \city{City}, \postcode{10587}, \state{State}, \country{Country}}}

%\affil[3]{\orgdiv{Department}, \orgname{Organization}, \orgaddress{\street{Street}, \city{City}, \postcode{610101}, \state{State}, \country{Country}}}


\abstract{
The body morphology plays an important role in the way information is perceived and processed by an agent.
We address an information theory (IT) account on how the precision of sensors, the accuracy of motors, their placement, the body geometry, shape the information structure in robots and computational codes.
As an original idea, we envision the robot's body as a physical communication channel through which information is conveyed, in and out, despite intrinsic noise and material limitations. %despite the constraints and uncertainties inherent in physical devices. %
Following this, entropy, a measure of information and uncertainty, can be used to maximize the efficiency of robot design and of algorithmic codes \emph{per se}.
This is known as the principle of Entropy Maximization (PEM) introduced in biology by Barlow in 1969. % to describe the efficient encoding performed by the brain.
The Shannon's source coding theorem provides then a framework to compare different types of bodies in terms of sensorimotor information.
In line with PME, we introduce a special class of efficient codes used in IT that reached the Shannon limits in terms of information capacity for error correction and robustness against noise, and parsimony.
These efficient codes, which exploit insightfully quantization and randomness, permit to deal with uncertainty, redundancy and compacity. These features can be used for perception and control in intelligent systems.
In various examples and closing discussions, we reflect on the broader implications of our framework that we called \emph{Informational Embodiment} to motor theory and bio-inspired robotics, touching upon concepts like motor synergies, reservoir computing, and morphological computation. These insights can contribute to a deeper understanding of how information theory intersects with the embodiment of intelligence in both natural and artificial systems.

}




\maketitle

\section{Introduction}
\label{introduction}

Before Claude Shannon's work, engineers believed that reducing communication errors necessitated increased transmission power or repetitive message transmission. Shannon fundamentally demonstrated that it wasn't essential to expend excessive energy, effort and time if appropriate coding schemes were employed~\cite{shannon_mathematical_1948, guizzo_closing_2004}, even in the face of an unreliable communication channel. We think that similar principles can be applied in robotics to enhance efficiency in perception, control, and body design. %despite the inherent unreliability of the body.
Information theory (IT) can provide insights to robotics in terms of efficient codes, information structure and communication between the body (hardware) and the controller (software).

\begin{figure*}[ht]
\vskip 0.2in
\begin{center}
\centerline{\includegraphics[width=14cm]{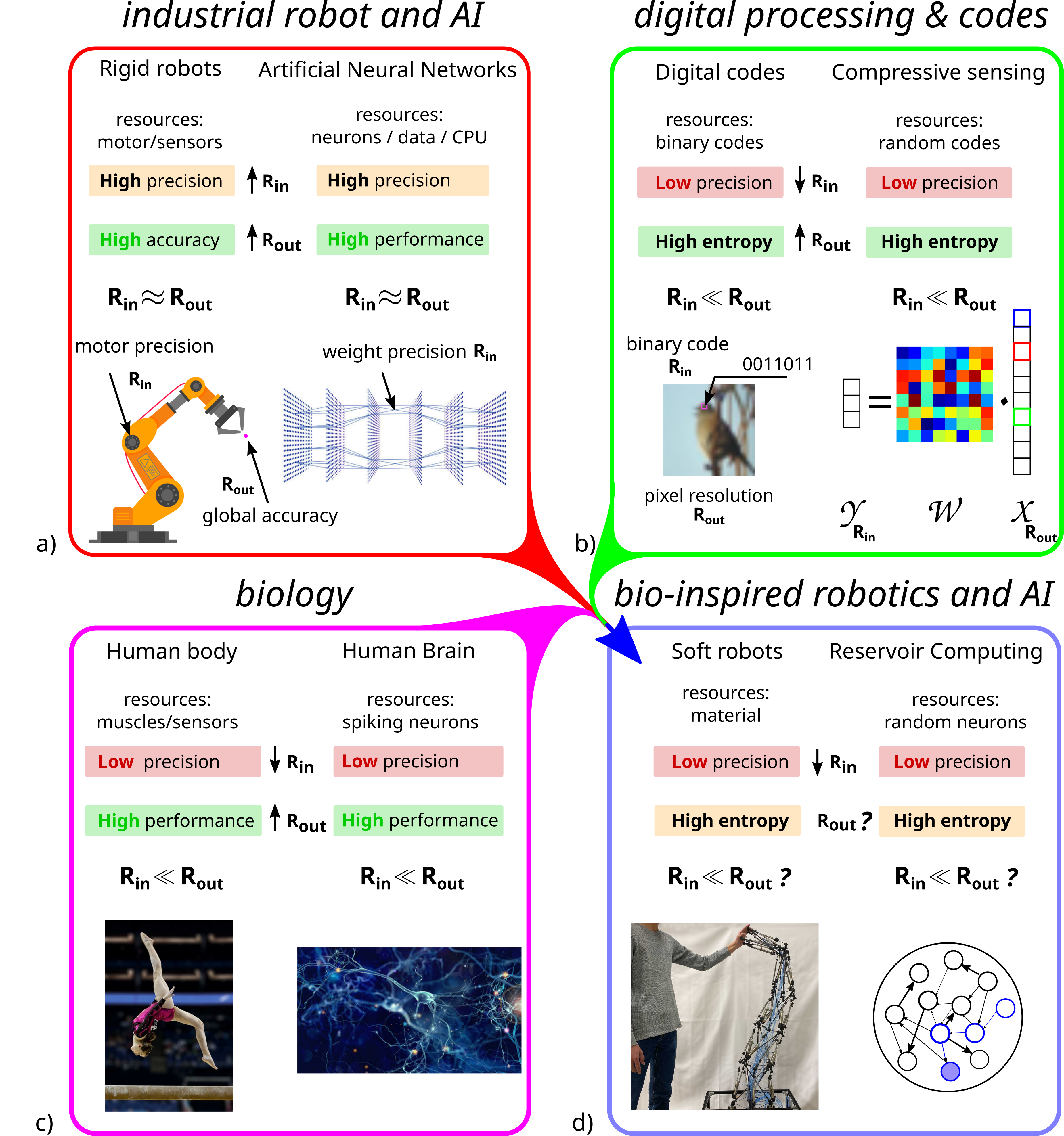}}
\caption{
Different strategies of information transfert in robotic and biological agents. a) rigid robots use very precise motors and sensors for perception and motion control. The accuracy of the motion behavior, $\bf R_{out}$ is related directly to the precision of the precision of the devices $\bf R_{in}$ so that $\bf R_{in} \approx R_{out}$. Gigantic models in current Artificial Intellgience (AI) uses the same strategy to obtain high performance. b) digital processing, instead, uses a different strategy with very raw binary codes to encode high resolution signals so that $\bf R_{in} \ll R_{out}$. Novel techniques from Information theory, such as compressive sensing, turbo-codes, or low density parity check, uses random matrices (high entropic codes) to achieve near Shannon limits in terms of information compression and robustness against noise. c) similarly, the human body and brain possess low precision muscles and neurons but achieve high performance so that we have also $\bf R_{in} \ll R_{out}$. d) Biologically-inspired robotics and AI systems may achieve high accuracy from low precision neurons and body by using similar techniques from Information Theory and observed in biology.}

%Information pipeline into a robot. The robot as a communication channel. Each layer computes and transmit information. Efficiency of the sensors as well as the codes have therefore a direct impact on the whole information pipeline. Besides, high precision devices and heavy computation controllers are not efficient as well in terms of energy consumption.

\label{informational_embodiment}
\end{center}
\vskip -0.2in
\end{figure*}

Current robots achieve high accuracy in motion control based on highly precise sensors and motors.
For instance, current electronics permit to control the motion behavior of robots at the speed of kHz with high resolution sensors to detect errors, simulate several actions and generate an optimal plan with precise motors.
Nonetheless, they rely heavily on high performance devices to work.
%They require nonetheless high speed processors for models' computation and decision making.
Therefore, the robot's accuracy in terms of performance $ \bf R_{out}$ is mostly linked to the precision level of its devices $ \bf R_{in}$, so that $ \bf R_{in} \approx R_{out}$ or even $ \bf R_{in} \gg R_{out}$.
In a sense, more precision in devices provides more accuracy in behavior, see Fig.~\ref{informational_embodiment} a).
% A similar issue occurs in current Artificial Intelligence (AI), in which massive data and computational resources ($\bf R_{in}$) are needed to achieve extremely high performances in cognitive tasks, $\bf R_{out}$. %either by parametric solving or Artificial Intelligence (AI)

In comparison, biological systems don't possess very precise sensors and actuators. However, they are capable to overcome their own limitations and reach higher accuracy in performance, so that $\bf R_{in} \ll R_{out}$. In a sense, they do more with less, see Fig.~\ref{informational_embodiment} c). Animals and current robotics use therefore different strategies to achieve efficient control.
Besides, contemporary robots still fall short of replicating the smooth and effortless variety and adaptive flexibility displayed by animals and average 18-month-old during everyday activities~\cite{pfeifer_challenges_2012, lerner_motor_2015}. %
%highly entropic systems with large combinatorics such as soft robots and reservoir computing (random networks). %This is because they would correspond
In return, bio-inspired robotics and neuro-inspired AI systems, such as soft robots and reservoir computing, may use in the future the same strategies as humans for motion control, in order to achieve high performance based on unreliable devices (e.g. sensors, motors, or neurons), so that $\bf R_{in} \ll R_{out}$, see Fig.~\ref{informational_embodiment} d). %. This is because they may develop large combinatorics as high entropic systems
In this line, we propose that IT can give insights how biological systems may use efficient coding to do it and how robots can reproduce it.

% In line with IT, Horace Barlow proposed in the 60' the principle of Entropy Maximization (EM) for biology~\cite{barlow_possible_1961, barlow_redundancy_2001}. The EM principle expresses that animals (brains) may maximize their informational resources (neurons) by using efficient codes to process information.
In the 60', Horace Barlow introduced the Efficient Coding theory in Biology~\cite{barlow_possible_1961, barlow_redundancy_2001}, suggesting that biological systems rely on an efficient coding mechanism to convey maximum information per unit~\cite{burdet_sensory_1999, olsson_sensor_2005}.
This theory, inspired by Jaynes' Principle of Entropy Maximization (EM) ~\cite{jaynes_information_1957, jaynes_prior_1968} and IT of Claude Shannon, posits that by reducing redundancy in a signal, we eliminate unnecessary predictable information, enabling storage of more information in a compressed code. Efficient codes are highly informative, compact, and robust over time, aligning closely with the Bayesian treatment of information.
Digital codes and current advanced methods in IT that will be presented in section 2 verify the Entropy Maximization principle so that we have also $\bf R_{in} \ll R_{out}$, see Fig.~\ref{informational_embodiment} b).
Accordingly, efficient coding can be advantageously used by animals for leveraging accuracy, in sensing, control, and information retrieval~\cite{laughlin_communication_2003}. % at the body level. %conveying the same idea of Shannon of efficient codes the brain may use efficient codes to maximize its neural resources for conveying maximal information,
%

%Applied to robotics, the Efficient Coding hypothesis provides insights into evaluating different bodies in terms of information processing and designing a body that maximizes information for sensing and control despite unreliable devices.

%Taken literally
Transposed into the perspective of embodiment, the body can be conceptualized as a resource, functioning metaphorically as a physical communication channel that conveys information in and out. Its characteristics in terms of information exchange, capacity and bandwidth will depend then on information conveyed by its morphology, its sensors, and its motors; which means its information structure. %,-- to which biological systems may exploit gradually the information capacity, and its bandwidth, to its absolute limits.
This view of embodiment is in line with the concept of morphological computation proposed by Rolf Pfeifer and colleagues~\cite{pfeifer_self-organization_2007, laschi_lessons_2016, muller_what_2017}, making explicit the link between the robot's body and computation, such that the robot's design has an incidence on the information structure and codes.
%Hence, motion control may be seen as the channel transmission between one intended act to its execution, as illustrated in Fig.~\ref{information_pipeline}.
%the morphological computation proposed by Rolf Pfeifer and colleagues.

From this perspective, effectiveness of one embodiment can be quantified in terms of information exchange on one side and in terms of information storage or encoding on the other side. Entropy, a measure for information, redundancy, and variability across systems, can assess then the physical limit of embodiment between the robot's body and its algorithm.
We propose that entropy can quantify the respective performances of different types of code and body, their bandwidth or their information bottleneck. % of the information plays a crucial role in Information Theory. %It is employed to dimension physical communication channels, assess the performance of codes, and define their absolute limits.

For the robotic engineer, these concepts borrowed from IT will help to answer practical questions such as: where to place the motors and the sensors in the robots in order to increase accuracy? What is an efficient controller for a specific body? Reversely, what is the limit in terms of accuracy of one specific body? What is its nominal speed and bandwidth? %And what codes can improve the robot's performance? %
And how many sensors or motors do I need for a certain precision level depending on their resolution and their placement?
Currently, these questions are solved only empirically based on experience by roboticists. IT can be a solid ally to robotics to conceptualize an \emph{informational} embodiment theory: an information-based theory to embodied agents and robots. %, informational an information-based theory to embodiment and robotics.

The paper is organized as follows. In section 2, we will present how efficient coding has been introduced in Telecom and Sensing Technologies in the 2000' and gave rise to an IT revolution. We will describe how these codes maximize entropy and how they can inspire embodied AI and robotics with novel concepts. % borrowed from IT such as redundancy, discreteness, randomness, resolution, which can be unintuitive.
% for efficient sensing, control and body design. For instance, the EM principle can provide to robotics novel insights and
%, applied to  sensors, motors, and morphologies.
%the link between How these codes relate to the Entropy Maximization principle, and how this principle can provide insights to robotics for efficient sensing, control and body design with respect to the resolution of sensors, and motors, and morphologies.
%
% In section 3, we will describe how the EM principle can give grounds to a well-known theory of embodied intelligence in robotics, the morphological computation proposed by Rolf Pfeifer and colleagues.
% As one unconventional consequence and paradoxical example, the EM principle may predict the long-term superiority of highly entropic systems with large combinatorics such as soft robots and reservoir computing (random networks). %This is because they would correspond to higher entropic systems with larger combinatorics.
%Because , and to a popular type of embodocontroller physical reservoir computing, a theory. for AI in comparison to current robotics and AI systems if properly designed for information processing

% Then,
In section 3,
we will present different examples supporting our informational embodiment theory. We will explain our framework in the light of quantification of different types of embodiment, of evaluating motor equivalence and information complexity, and the effect of the codes' structure on sensory fusion. %This will be done by using concepts from IT such as redundancy reduction, combinatorics, number of degrees of freedom, quantization, dense coding, uncertainty.
This section will end by presenting links with current motor theories in human control, and acquisition of motor synergies during development. % information quantification of different behaviors based on their combinatorics.

We will discuss then pros and cons of our proposal and its links with other frameworks toward the understanding of biological intelligence and motor development in humans, supporting the contribution of embodied intelligence to current robotics and AI.

% % FIG 1
% \begin{figure*}[ht]
% \vskip 0.2in
% \begin{center}
% \centerline{\includegraphics[width=16cm]{images/information_pipeline11.png}}
% \caption{Information pipeline into a robot. The robot as a communication channel. Each layer computes and transmit information. Efficiency of the sensors as well as the codes have therefore a direct impact on the whole information pipeline. Besides, high precision devices and heavy computation controllers are not efficient as well in terms of energy consumption.}
% \label{information_pipeline}
% \end{center}
% \vskip -0.2in
% \end{figure*}

%\section{Principles from Information theory, Maximum Entropy, and Efficient Codes}

%
% {\bf The informational body. }
\section{Efficient Codes, the Body as an Information Channel. }

Our hypothesis draws inspiration from the Embodied Intelligence (EI) paradigm developed by~\cite{steels_intelligence_2018, brooks_building_1993, pfeifer_self-organization_2007, pfeifer_cognition_2014,nakajima_information_2015, muller_what_2017} and others that explicitly delve into information theory~\cite{hutchison_information-theoretical_2004, lungarella_how_2007, lungarella_information_2007, thornton_gauging_2010}. While EI has provided essential principles, it is still in search of a comprehensive theory~\cite{zahedi_quantifying_2013, ghazi-zahedi_morphological_2019, mengaldo_concise_2022}.

% One important concept, Morphological Computation, makes explicit the link between the robot's body and computation, such that the robot's design has an incidence on the information structure and codes.
%
We propose that efficient coding --, in the sense given by Shannon of entropy maximization,-- can explicitly describe the design of sensors, controllers or bodies that maximizes information processing.
% In contrast, the informational aspect of the robotic body, which is the most important part, is completely unthought by the engineers. For the roboticists, the design of the robot's morphology, its geometry, the correct placement of the sensors and of the motors are thought mostly from a control theory viewpoint only, and in effect, they are mostly a matter of experience and of rule of thumbs to achieve good information processing within. %For instance, it is the charge of the engineers to design properly
%

% However, the body design has an incidence on the information structure. It requires the careful placement of the sensors and of the motors, and their precision level. These characteristics have a consequence on the shape of the information structure and the efficiency of the controller, the learning system and the sampling of the environment.

{\bf Efficient Coding in Biology. }
Efficiency coding principles have been proposed in Biology in which original signals, redundant, are hypothesized to be transformed into neural codes of uniform distribution~\cite{laughlin_simple_1981, van_hateren_theory_1992}. By doing so, codes are more compact and can transmit the same amount of information with far less channel capacity~\cite{van_hateren_theory_1992, atick_what_1992}; hence maximizing information.

For example, the human optical nerve, which is composed of 1 million ganglions, has to transmit the information coming from the retina, which is composed of 120 millions retinal photo-receptors.
Following the hypothesis of efficient encoding that neurons should encode information as efficiently as possible in order to maximize neural resources, it has been shown that visual data can be compressed up to 20 fold without noticeable information loss.
The first experiences in testing the theory of efficiency coding or redundancy reduction in Biology came from the work of Laughlin~\cite{laughlin_simple_1981} applied to the fly eye. He measured and compared both the contrast distribution in the image and the contrastive cells in the fly eye and predicted that optimal encoding would take the form of maximizing contrast by transforming the original (redundant) distribution into a uniform (uncorrelated) distribution to be transmitted to the fly brain.
As each output value becomes equiprobable, the conveyed signal achieves the capacity limit for transmission with optimal bandwidth~\cite{laughlin_communication_2003}. As a result, optimal coding makes the signal resemble white noise (Maximum Entropy): a coding effect that is called whitening~\cite{van_hateren_theory_1992}.

%
% Another example of efficient coding is observed in the optic nerve, which constitutes a bottleneck in transmitting information to the brain as it comprises 1.7 million ganglion cells, although the number of photoreceptors is on the order of 126 million cells. A reduced code constructed from the difference between photoreceptors in the retina (e.g., a differential code) is sent to the brain with the same amount of information, which requires far less channel capacity~\cite{van_hateren_theory_1992, atick_what_1992}.

%
%Hence,
Efficient coding is hypothesized to occur widely in the brain to manipulate natural input, and EM appears to be one important principle of brain dynamics~\cite{jirsa_entropy_2022, luppi_information_2024, kringelbach_thermodynamics_2024}. The level of redundancy in each brain region is suggested to relate then to different types of computation and treatment of information; e.g. for memory access, storage, and retrieval~\cite{rolls_neuronal_2011, rolls_pattern_2016}. %been observed with respect to the brain area. %The same hypothesis is expected for memory access, storage, and retrieval in areas such as the hippocampus and the prefrontal cortex (6, 13).

% In so far, the efficient coding hypothesis has been only marginaly applied to robot sensing and control, with some exceptions~\cite{olsson_sensor_2005}.
%

For instance,
the models of sparse coding and motor synergies, in human sensing and motor control, illustrate as well the efficient coding hypothesis with pattern separation and grouping in the human body~\cite{bernstein_coordination_1967, bizzi_combining_2008, de_rugy_are_2013, morasso_vexing_2022}.
%
% are very well known theories of sensing and motor control~\cite{bernstein_coordination_1967, bizzi_combining_2008, de_rugy_are_2013, morasso_vexing_2022}. In our framework, they can be viewed as efficient codes in terms of information compression.
%
%
Motor synergies correspond then to the dynamical combination of a discrete set of motor groups to realize motion behaviors. They are encoded at the level of the spinal cords in the place of central pattern generators for the rhythmic patterns and of reflex circuits for those driven by sensory feedback. Since their number is limited in comparison to the number of muscles to control, they can be viewed then as sort of ``compressive codes'' to convey information.

%In comparison with the previous examples on sensor encoding,
% For instance
In this line, the spinal cords system can be seen as a communication bottleneck in terms of motor access for the brain since the ratio between the number of these spinal cord cells and the number of muscles is very low~\cite{morasso_vexing_2022}.
As a proposal, the efficient coding hypothesis can serve to design a reduced number of motor codes or synergies optimized for control, adapted to one particular body, yielding to minimal energetic cost of transport and best stability. %,-- has not been clearly proposed before.
%
% In this line, some new research support that optimization in the presence of uncertainty can explain how the circuits should best be combined for locomotion~\cite{ryu_optimality_2021}.
%
% We suggest instead that uncertainty, and entropy, has to be taken into account at the motors and sensors level as it has an incidence on the way these units are encoded to maximize information exchange, or control, during interaction with the environment.

% FIG 2 EFFICIENT CODES
\begin{figure*}[ht]
\vskip 0.2in
\begin{center}
\centerline{\includegraphics[width=16cm]{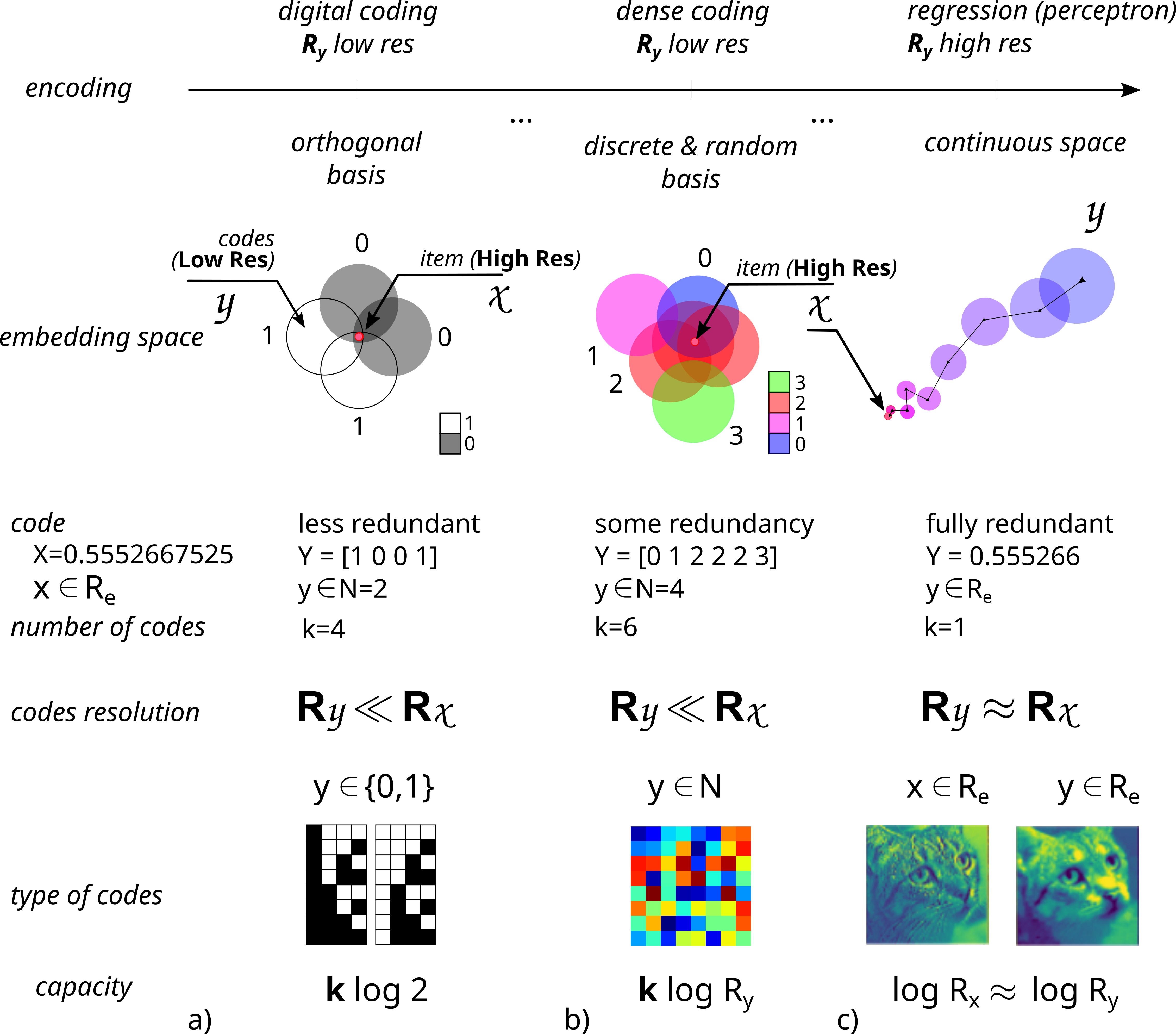}}
\caption{Information structure for different types of codes. a) Digital encoding represents directly a high resolution input $\bf X$ with orthogonal and low resolution (binary) codes $\bf Y$ so that the number $\bf k$ of codes is the shortest, with $\bf R_Y \ll R_X$. b) random codes used in Telecom and signal exploit a similar strategy to digital processing by creating orthogonal embeddings of low resolution on the fly so that $\bf R_Y \ll R_X$. The number of codes $k$ is small so that $\bf \log R_X \approx k \log R_Y$. c) Besides, highly redundant codes require high resolution devices to approximate the input values with same resolution, so that $\bf R_Y \approx R_X$.}

% Compressive methods in Telecom and signal have delivered efficient methods for compression and redundancy reduction using 'simple' random matrices to collect and compress at the same time information. Random projections of one original signal --, and also permutations,-- don't destroy information but create orthogonal embeddings readily. Similar with sparse coding in biology, these random projections represent then some short codes of low dimension, which can serve to detect the primitives in high dimensional signals or reconstruct them back; results taken from~\cite{pitti_digital_2022}.

\label{structure_codes}
\end{center}
\vskip -0.2in
\end{figure*}

{\bf Randomness, redundancy reduction and compressive codes. }

\begin{figure*}[ht]
\vskip 0.2in
\begin{center}
\centerline{\includegraphics[width=16cm]{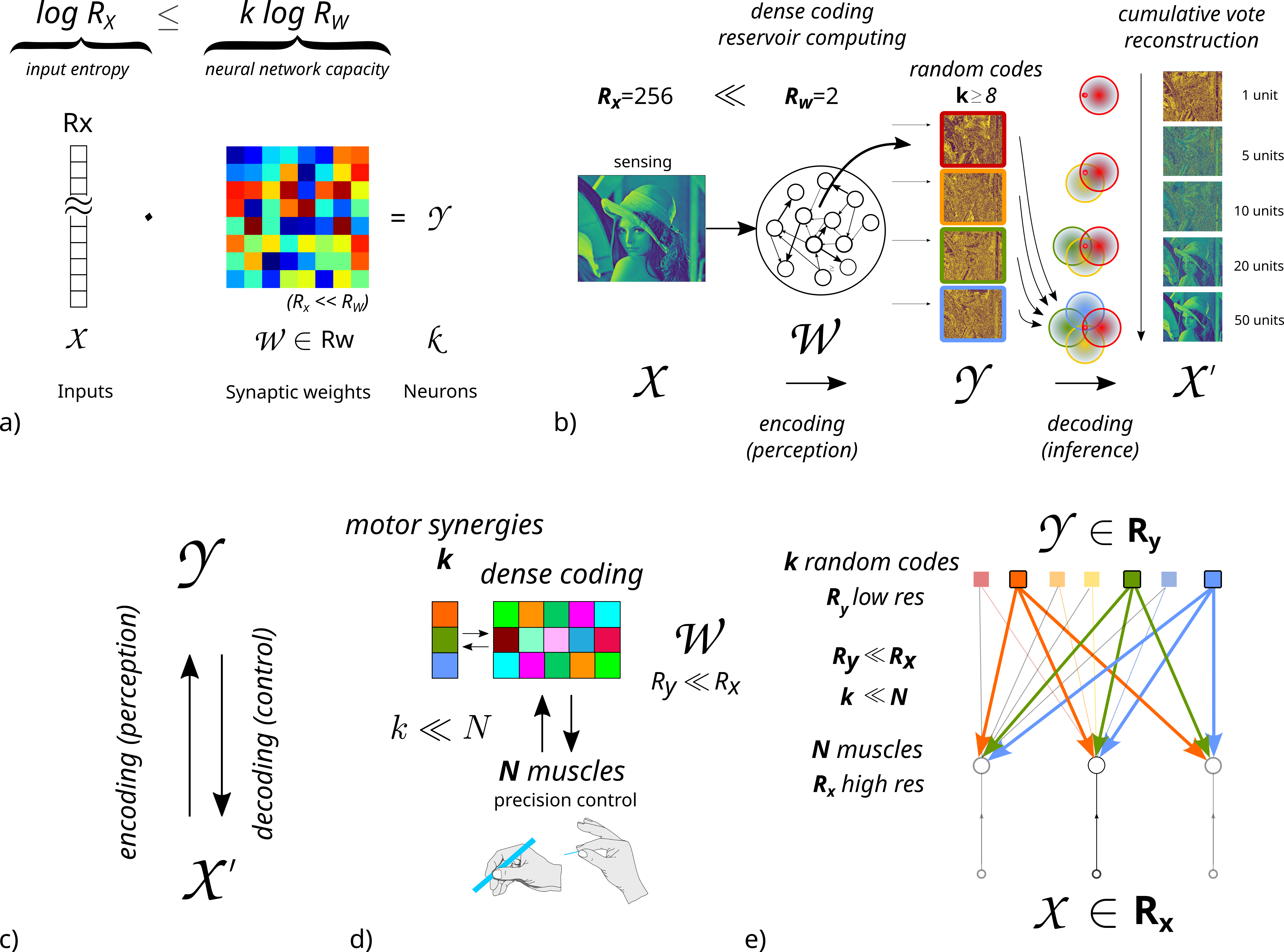}}
\caption{Efficient coding in random matrices in Compressive sensing, application to sensing and control in robotics. a) Compressive methods in Telecom and signal have delivered efficient methods for compression and redundancy reduction using  random matrices $\bf W$ to collect and compress at the same time high dimensional signals $\bf X$ into only few codes $\bf Y$ so that $\bf k \log Y \approx \log X$, with $\bf k$ the number of units. b) Random projections of one original signal --, and also permutations,-- don't destroy information but create orthogonal embeddings readily. Information is compactly preserved into few codes that can reconstruct back information by inference (decoding); results taken from~\cite{pitti_digital_2022}. c) efficient codes can serve to compactly regenerate high dimensional signals into a lower dimensional space. d-e) In motor domain, they represent then motor primitives or motor synergies.}
\label{sparse_coding}
\end{center}
\vskip -0.2in
\end{figure*}

% 1 EM in signal and turbocodes
%Telecom and Sensing Technologies can provide some answers.
Twenty years ago, an IT revolution occurred, introducing methods that exploited advantageously the EM principle with the use of random matrices to transmit and encode information near error-free, see Fig.~\ref{sparse_coding} a). The random nature of these codes $ \bf y$ turned out to be critical to correct signals and messages $ \bf x$. Accordingly, although these codes $\bf y$ can be highly corrupted by noise or purposefully limited by low resolution $ \bf R_{y}$, it turns out that a very small number $\bf k$ of units are sufficient to approximate signals $\bf x$ of very high resolution $ \bf R_{x}$; so that $ \bf R_{y} \ll R_{x}$. $\bf R_y $ and $\bf R_x$ correspond respectively to $\bf R_{in}$ and $\bf R_{out}$ of Fig.~\ref{informational_embodiment}.
These random codes were the first to reach the Shannon limit in terms of transmission. As such, they were also the first to satisfy the source coding theorem: $ \bf \log R_{x} \approx k \log R_{y}$ with $ \bf R_{y} \ll R_{x}$. These codes, and the principles behind, can provide therefore hints on how reliable and efficient robot's sensing, control and body design can be, despite unreliable sensors, motors and body morphologies.

% Twenty years ago, a significant IT revolution occurred in Telecom and Sensing Technologies, introducing methods that exploit the Maximum Entropy Principle through the use of random matrices.
Turbo-Codes~\cite{berrou_near_1993, guizzo_closing_2004}, Compressive sensing~\cite{candes_stable_2005, donoho_compressed_2006}, and Low-Parity Density Check (LDPC) [ref], which are concrete applications, enable fast and accurate sensing using random matrices and permutations. Random projections of an original signal, convoluted by dense random matrices, create orthogonal embeddings that minimize the necessary sampling for encoding, achieving near-Shannon limits for transmission.

In comparison to standard methods that used $\bf N$ samples for encoding one signal, compressive sensing requires only $\bf k$ samples of low resolution $\bf R_y$ to reconstruct back the image of higher resolution $\bf R_x$ such that $\bf R_y \ll R_x$, and $\bf k \ll N$. A schematic example is presented in Fig.~\ref{structure_codes} b) to illustrate its functioning in which random patches $\bf y$ intersect each other to represent one item $\bf x$ so that $\bf R_y \ll R_x$. Although each patch cannot represent precisely the item due to their low resolution, the combination of few are enough to recompose it.

%Compressive sensing with random codes can encode signals directly with low resolution patches.
In comparison, regression methods such as the descent gradient used in artificial neural networks (perceptrons) or Free-Energy Minimization~\cite{friston_free_2006, pitti_iterative_2017, pitti_brain-inspired_2021} require longer time for convergence by removing iteratively
errors to approximate the signal, see Fig.~\ref{structure_codes} c). At the end of the iterative stage, the synaptic weights $ \bf W$ approximate the signal $\bf x$ so that the resolution of the synaptic weights $ \bf R_W$ has the same resolution as the signal $\bf R_x$, or higher such that $ \bf R_W \approx R_x$. As a result, these methods require very precise devices to work.

Importantly, these results demonstrate that information structure plays a computational role for capacity storage, and robustness against noise.
For instance, while discrete and non-redundant codes can be stored and retrieved very rapidly, analog and redundant codes require more precision or capacity storage. Counter-intuitively, very precise codes are also more prone to noise because they rely on very precise devices in counter part.
In comparison, discrete codes, which remove redundancy in the signal, makes them also more robust against variabilities. They are therefore easier to handle for information preservation and retrieval.

In computers and modern communication systems, the digital treatment of information proposed by von Neuman and Claude Shannon using binary codes (low resolution) instantiates fully this efficient encoding principle. Because information is encoded using a limited set of discrete values, only binary values, digital encoding instantiates Entropy Maximization, and satisfies the source coding theorem such that $\bf \log R_x \approx k \log R_y$, with smallest number $\bf k$ of codes; see Fig.~\ref{structure_codes} a).
It follows that each bit added to the codeword augments the resolution by a power-law factor as it is known.

%Interestingly, in human communication systems, maximizing information or entropy in one channel corresponds to a digital treatment of information in which each bit added to a codeword augments the resolution by a power-law factor.
%
% Therefore, because the type of processing performed is about maximizing information, we suggest that this neural network is one of the first instance of neural networks that we can term digital.
%
% While robotics do use digital processing for sensors and motors encoding, and algorithmic programming, the maximization of entropy or information is not used as a design principle.
% Althought the brain is not a computer, embodied Intelligence may be digital in the sense given by Shannon, .

% FIG 3 COMPRESSIVE SENSING

% FIG 4 MOTOR EQUIVALENCE
\begin{figure*}[ht!]
\vskip 0.2in
\begin{center}
\centerline{\includegraphics[width=16cm]{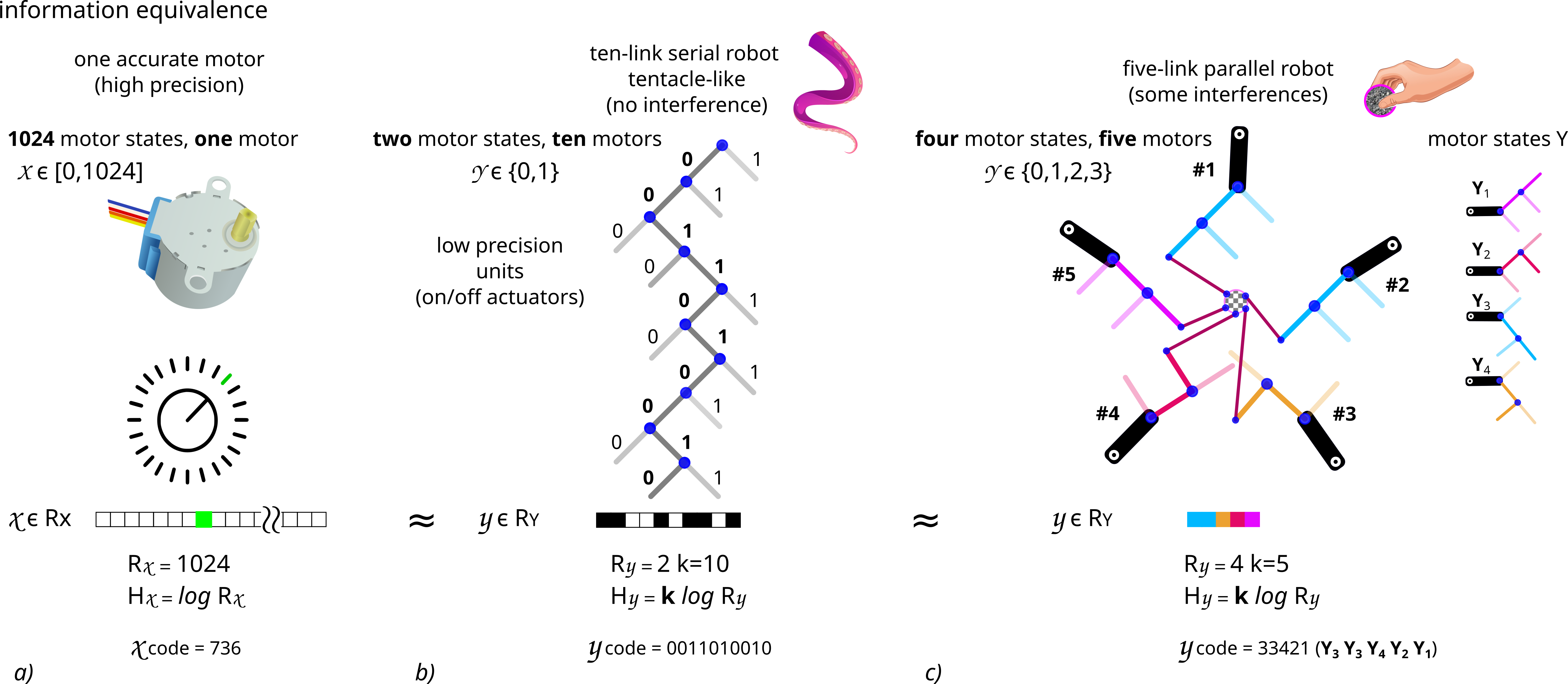}}
\caption{Information equivalence of different embodied systems. One motor device $X$ of high precision within $R_X$ is equivalent in term of information to the combination of multiple actuators $Y$ in series of lower dimensions, say $R_Y$, resp. a) and b). In the case where $R_X=1024$ and $R_Y=2$ (on-off actuators), one equivalence exists in terms of information between the two systems for a minimum number of $k= \log R_X / \log R_Y = 10$ units. c) same equivalence for a hand-like five-link parallel robot with $4$ possible motor states. The combinatorics is also $1024$. Image credit upklyak, ddraw.}
\label{equivalence_embodiment}
\end{center}
\vskip -0.2in
\end{figure*}

{\bf EM in random neural networks. }
%
%We believe that this interpretation of Shannon's source and channel coding equations is innovative and can significantly influence the way we design robots and controllers in the future.
In this line, the EM principle was applied in sensory encoding for maximizing information~\cite{burdet_sensory_1999, olsson_sensor_2005} and in artificial neural networks for efficient encoding within random networks~\cite{pitti_digital_2022}.
%
% In the later, the objectives were to replicate the kind of digital processing used by the brain to support EM. % with simple mechanisms.
%
To do so, the mechanisms presented above of discreteness and randomness were used to remove redundancy in the signal and to create orthogonal neural codes, see Fig.~\ref{structure_codes} b) and  Fig.~\ref{sparse_coding} a).
%In~\cite{pitti_digital_2022}, we hypothesized that the brain supports EM by performing a kind of digital computation of information by using such simple mechanisms at the neural level.
% In this paper, we believe that these design principles can be expanded to embodied AI and bio-inspired robotics as well for the maximization of information.
% althought the brain is not a computer,
%~\cite{pitti_digital_2022}. We modeled this principle in an artificial neural network

Results in~\cite{pitti_digital_2022} demonstrated that the neural network reaches Shannon's capacity limit in terms of information storage per neuron within it, see Fig.~\ref{sparse_coding} b). They showed that random neurons can convey maximal information, so that each neuron added to the neural population increases its storage capacity by an exponential scale, like digital processing. As a result, few neurons $\bf y$ are enough to encode one signal $\bf x$ of high resolution so that we have effectively $\bf R_y \ll R_x$.

Because this neural network verifies the source coding theorem with $\bf \log R_x \approx k \log R_y$, it performs a kind of digital computing, in the sense given by Claude Shannon of Entropy Maximization.
We expect that these results and their mechanisms can be transposed to embodied AI and robotics.
For instance, random networks can be used to maximize information either for fast sampling in few shots (encoding) or for motor control with few primitives (decoding), see Fig.~\ref{sparse_coding} c-e).
% as well for the maximization of information.
% While robotics do use digital processing in sensors, motors, and controllers, the EM principle is not used as a design paradigm to provide efficient codes as biological systems do to reach high accuracy in performance.
% Althought the brain is not a computer, embodied Intelligence may be digital in the sense given by Shannon, .

At the conceptual level, it suggests that the information structure of codes --i.e., their level of redundancy, discreteness, their resolution-- has an impact on their computational properties. Hence, poor sensors and unreliable motors (low entropic systems) may approximate accurate sensors and precise motors (higher entropic systems).
Accordingly, roboticists can play with the body structure, their combinatorics, their degree of redundancy, the resolution of their sensors, and motors to develop accurate robots, soft and rigid, that maximize information.

Thus, we suggest that the level of embodiment of one system (a robot or a biological agent) correlates with its capability to convey information~\cite{lungarella_mapping_2006, lungarella_information_2007, zahedi_quantifying_2013, ghazi-zahedi_morphological_2019}. Our approach based on EM is in line with two proposals of embodied AI, namely morphological computation and physical reservoir computing, that we will present in the Discussion section.

\section{Some examples}

{\bf Information equivalence for different robot bodies}. Let's use Shannon equation of source coding $\bf \log R_X = k\,\log R_Y$ applied to a simple robotic example to describe its use, see Fig.~\ref{equivalence_embodiment}.
Let's have a very precise motor or sensor, $\bf X$, such that it can take any values within the interval range $\bf R_{X}$; its resolution.
For the purpose of our demonstration, let's assume that $\bf R_{X}=1024$.
Accurate sensing or control with such device is difficult since it requires the same amount of precision for its calibration by another sensor or motor.

According to Shannon equation applied to robotics, one such high resolution device is equivalent in terms of entropy to $\bf k$ low resolution devices $\bf Y$; $\bf k$ being the number of devices.
Let's assume two other robotic designs in which we change the number of actuators, their resolution, and their disposition; resp. Fig.~\ref{equivalence_embodiment} b) and c).
For one robotic design similar to a kinematic chain, we select the number of states of the devices $\bf Y$ to have only two states so that $\bf R_{Y}=2$; i.e., such system may correspond to agonist and antagonist actuators, or on and off sensors.
At a biomechanical level, this design resembles to a tentacle.
%For the purpose of our demonstration, let's assume that the number of states of the devices $Y$ are only of two states so that $R_{Y}=2$; i.e., such system may correspond to agonist and antagonist actuators, or on and off sensors.
%
In comparison to the first device, this second one is simpler to control as each motor possesses two states only, which are easier to discriminate than high precision motor.
As a kinematic chain, each motor does not affect the state of another. This system has therefore no redundancy.

The third morphology corresponds to a five-link redundant robot with four-state actuators; see Fig.~\ref{equivalence_embodiment} c).
If the design is fully rigid, there would be some impossible states. So let's assume some flexible wires, in purple, so that all the states are possible.
This morphology ressembles roughly the one of the human hand with its five fingers manipulating the same object.

According to Shannon inequality equation of source coding between the three systems, we will have correspondence when $bf \log R_X = k\,\log R_Y$.
Thus, one equivalence from an IT viewpoint between them will occur for $\bf k = \,\log R_X / \log R_Y = 10$ and $\bf k \approx 5$, respectively for the second and third architecture; $k$ being the number of necessary low resolution devices $\bf Y$ to approximate the device $\bf X$.
%
%The actuators have just two more states than those of the previous architecture.
In comparison to the first system with one accurate motor, the two last architectures have very few motor states. However, they possess the same number of states and therefore the same complexity level.

Therefore, in terms of accuracy and information, which are employed as synonyms here, one series of ten switch motors or a five-link parallel device with just two additional states (four states) worths one very precise motor of 1024 states.
This indicates that multi-articulated systems can gain in precision and accuracy even with simple and non precise controllers.
Furthermore, the relationship between precision and the number of motors is asymmetric in the equation $\bf \log R_X = k\,\log R_Y$. Precision increases following an exponential scale whereas the number of controllers $\bf k$ augments following a linear scale.

{\bf Information hierarchies in motor and sensor grouping}.

%tTing
% Dancers have rigorous training in agility and balance that affects not just their bodies, but the neural mechanisms of movement control. We have leveraged the specific training involved in different dance forms to test theories about training and experience alter how individuals move across the spectrum from motor skill to motor impairment.
% Certain aspects of motor skill learning in dance may transfer to other challenging balance tasks and shape how we walk in everyday life. We hypothesize that muscle coordination patterns–called motor modules or muscle synergies–are learned building blocks of movement that give rise to “motor accents”, or individual styles in movement. We showed that pre-professional and professional dancers use similar motor modules in a challenging beam traversing task (Sawers and Ting 2015) and in normal overground walking whereas non-dancers did not (Sawers et al 2017).  Thus, motor skill learning as in dance appears to shape existing movement patterns for everyday task. Motor skill learning in dance may also expand the capabilities of those patterns to accomplish more challenging motor tasks. Not only could generalizability of motor modules tasks play a role in why we may be recognized based on our gait, it also informs how dance-based and other rehabilitation training may also generalize to real-world situations.

% FIG 5 JOINT ACTIONS
\begin{figure*}[ht]
\vskip 0.2in
\begin{center}
\centerline{\includegraphics[width=16cm]{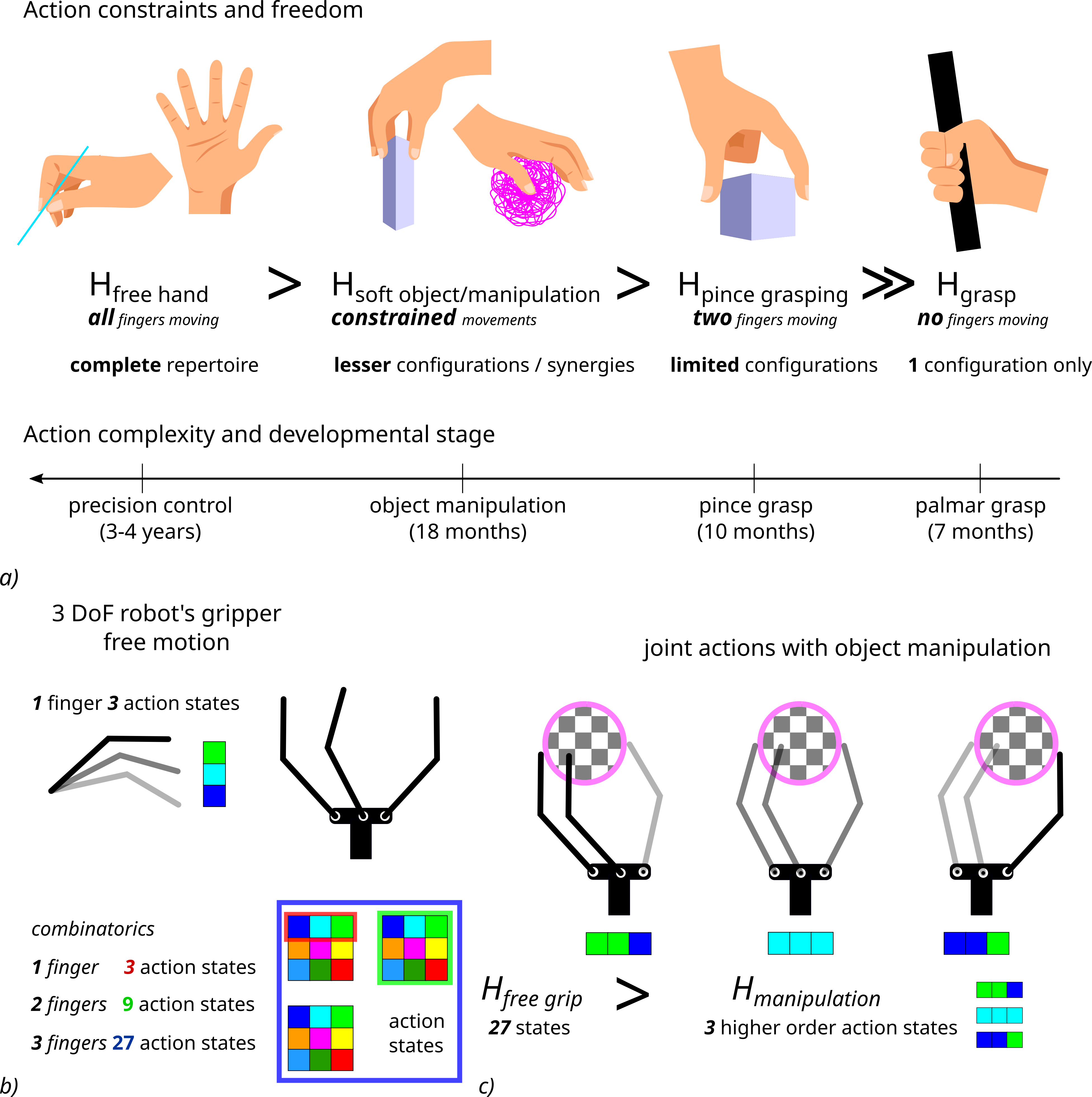}}
\caption{Combinatorics of one hand motion, quantification of contacts and constraints. a) Free hand motions are the most entropic ones $H_{free}$ as they are also the most unconstrainted ones. This state corresponds to the complete repertoire of actions. Besides, constrained actions occuring when touching soft and rigid objects can be quantified by the number of possible configurations $H_{soft} > H_{rigid}$. The full grasp of one object is paradoxally the most minimal and the less entropic one, as it corresponds to only one possible configuration of the hands b-c) same explanation for a more simplistic robotic hand with 2 DoFs and 9 possibles actions. The grasping of one object represents to a lower entropic state than one of a free-hand motion, and thus, to a more stable state. Image credit ddraw.}
\label{joint_action}
\end{center}
\vskip -0.2in
\end{figure*}

The example above in Fig.~\ref{equivalence_embodiment} presented the effects of motor grouping in terms of information gain depending on their organization (serial or parallel) and on their resolution (small or high precision).
The source coding theorem explains how this gain grows exponentially when redundancy is removed even for unreliable devices. % can become efficient when they work together.

A second property of this theorem is the notion of information hierarchy coming from the inequality between the left and right parts of the equation $\bf \log R_X \le k\,\log R_Y$.
For instance, we can chain at different organization level the capacity limit of units so that we can have $\bf j \log R_{X+2} \le k\,\log R_{X+1} \le l\,\log R_{X} \le m\,\log R_{X-1}$.

The coordination of individual motors, which means the sets of all possible groups, constitutes a combinatorics of $\bf k\,\log R_Y$.
% Coordination reduce at the same time redundancy and entropy at the level of longer motor sets $\log R_X$.
%Hence, there is a beneficial effect in terms of information $\log R_X$
If the entropy of longer motor sets (chunks) have a lower entropy of $\bf \log R_X$, then coordination constitutes a beneficial effect. Similarly, this property has been used  by Claude Shannon to describe the benefit of grouping letters into pairs, triplets and then into words in the English language for robustness against noise, faster encoding and better inference~\cite{shannon_prediction_1951}.
Because the number of words are lesser than the combinatorics of random letters, they save resources during encoding and energy  during retrieval. %  words combination of Smaller number means that we can pre-activate less brain areas for cognition to save energy.

Similarly in the motor domain, the motor synergies limit the variabilities of individual muscles with fewer coordinated patterns to control. Motor synergies are therefore compressive, and represent an economical gain.
Motor synergies for robotics and biological systems are therefore advantageous from an information gain perspective. They can be retrieved by the redundancies found in the body morphology; i.e., its symmetries.

{\bf Information equivalence in Sensory arrays}.
% Similarly
In the sensory domain, efficient coding will be realized in sensor arrays if they maximize information with respect to their number and their resolution.

For a sensory array $\bf X$, let's say a camera, with pixel resolution, $\bf R_x=256$.
This device $\bf X$ can be theoretically replaced by $\bf k=8$ distinctive cameras $\bf Y$ of very low resolution, $\bf R_y=2$ or by one moving agent possessing one camera $\bf R_y=2$, and sampling eigth times the object from different viewpoints, see Fig.\ref{sparse_coding} b). Following the criteria for digital encoding, the minimal number of low-range sensors required to approximate the higher resolution $\bf R_x$ is then $\bf k= \log R_x / \log R_y \approx 8$. This situation is comparable with the example presented in Fig.\ref{structure_codes} a). In the case of randomly placed low-range sensors, as presented schematically in Fig.\ref{structure_codes} b), slightly more cameras are needed to approximate the image resolution $\bf R_x$ as they will have some redundancies. However, this second approach presents some robustness against units failure. For example, having multiple cameras is less critical to rely on in comparison with only a unique high resolution camera that could fail, see Fig.\ref{structure_codes} c).

% Besides, Random binary neurons as presented in Fig.\ref{structure_codes} b) and in Fig.\ref{sparse_coding} can still approximate this resolution but with a little more neurons, see~\cite{pitti_digital_2022}.

% This result has been already demonstrated in CS where a one-dot camera has served for the reconstruction of a much higher resolution image despite the ambiant noise [ref].
%
These results express that interaction between devices, even of low resolution, can rapidly improve the information gain following a power-law scale. The benefit can be important even with a small number of devices. Low-range sensors benefit from the random projections to create dense orthogonal representations. For instance, a one-dot camera has been used for the reconstruction of a much higher image resolution, and despite the ambiant noise, by using a random grid [ref]. %Thus, this strategy introduced in CS differs from conventional encoding methods found in robotics.

{\bf Information gain through mutualization}. What is the information gain or loss of mutualizing control and sensing instead of doing it separetedly?
%
% When the body moves, it imposes a constraint on the possible movements it can perform.
%
Even when it is said that the body moves 'freely' --e.g., the arms, the fingers, or the head,-- the body's morphology imposes some constraints on the possible movements it can perform. We propose that the number of those combinatoric possibilities corresponds to its entropy.
These are not the only constraints. Those coming from physical interaction with the environment or from different body parts are additional ones.

For instance, constrained motions occur when our fingers become virtually bound together when they manipulate one object, when an objet is moved with the two hands, or when we just stand upward with the two feet 'locked' on the ground.
One consequence is that when more constraints are imposed to the body, the number of states diminishes, and as a result, its entropy becomes lower.
Therefore, even if we interact with a soft object, a spounge, we can discriminate the number of possible configurations it corresponds, and quantify it with respect to the free hand motion.

Let's have one example of a robotic hand whose fingers can move independently and freely, see Fig.~\ref{joint_action} b).
In this case, the number of states and possible fingers' configuration is the largest and corresponds to the most entropic value of the hand $H_{free}$.
For a three finger's hand, and for each finger moving in three positions only, the number of possible states is $R_{free}= 3^3$, which is also the highest entropic value of the robotic hand.

% FIG 6 MOTOR UNCERTAINTY LEARNING AND ADAPTATION
\begin{figure*}[ht]
\vskip 0.2in
\begin{center}
\centerline{\includegraphics[width=16cm]{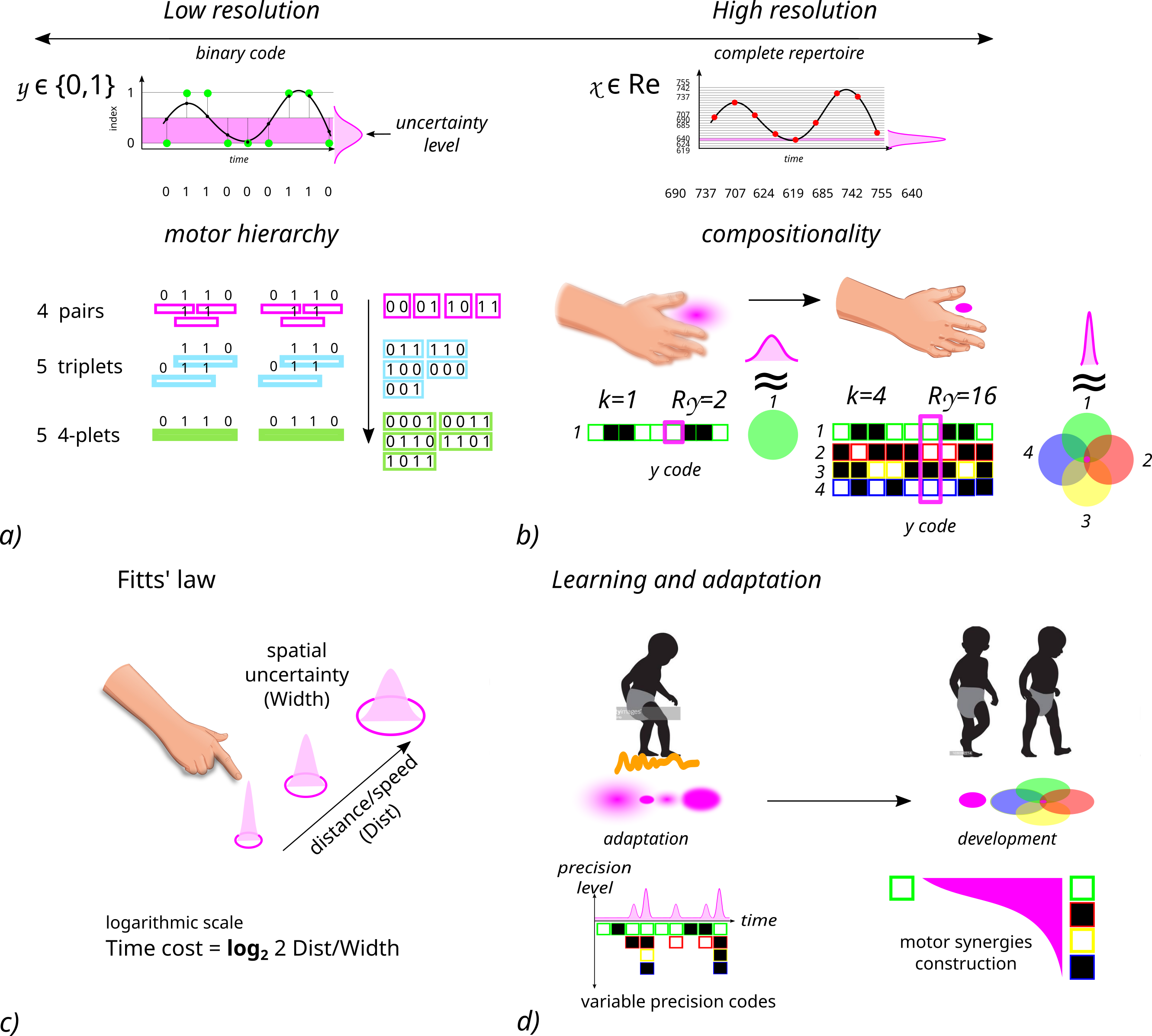}}
\caption{Motor uncertainties, adaptation and learning. a) Low resolution codes can approximate much higher-resolution signals in few incremental steps, either sensors or motors, according to Shannon's source coding theorem. The precision level in motion control can be leveraged by the number of sensorimotor codes, and their sparsity (difference) guaranties their small number. b) The Fitts' law in motion control describes the logarithm nature of motor accuracy, and its bandwidth, which is dependent to the distance to the target and the arm speed. c) Motor adaption can be described as aligning one's own motor accuracy to the environment's uncertainties. The number of recruited codes can be tuned to the desired stiffness level. d) Similarly, motor development can be seen as the learning of the specific number of motor primitives tuned to the correct level of the body's accuracy. Image by ddraw on Freepik. JakeOlimb GettyImages.}
\label{motor_resolution}
\end{center}
\vskip -0.2in
\end{figure*}

In comparison, object manipulation has for effect to constraint the number of possible states and therefore the entropic value of the hand, $H_{obj\_in\_hand}$, such that $H_{free} \gg H_{obj\_in\_hand}$.
In our case, we would have $R_{obj\_in\_hand}=3$ possible action states, see Fig.~\ref{joint_action} c). Therefore, the relative entropy between the two situations is a large decreasing of the possible action states to $R_{obj\_in\_hand}/R_{free}=3/27=1/9$. The entropy of the three fingers hand which manipulates one object becomes equivalent to the entropy of solely one free finger. The relative entropy loss is then $\log R_{obj\_in\_hand} - \log R_{free}=-0.95$.

If we transcribe this example to the human hand as in  Fig.~\ref{joint_action} a), we will have the highest entropic value associated to the situation of the free hand motion; e.g., $H_{free}=5 \log R_{finger}$ with $R_{finger}$ the number of states of one finger.
Now, if we manipulate a soft object, the number of possible actions will be still high but lower than the case of free hand motion.
Besides, if a rigid object is held by two fingers only like a pen, the number of configurations will drastically decrease.
Roughly, the information loss will be $H_{obj\_in\_hand} < 2 \log R_{finger}$.
Next, if we hold the object with three fingers this time, we will have a certain degree of freedom, but lower than the case of two fingers.
Finally, the full grasping of the rigid object enclosed by all the fingers is the most constraining case, of one state only; e.g., $R_{closed}=1$ and $H_{closed}=0$. It would correspond to the lowest entropy values in comparison with the case of manipulation by the five fingers for instance.

From an information viewpoint, it makes sense that the most constrained states are also the lowest entropic ones. It is also intuitive that constrained and mutualized actions should demand less cognitive charge and computational resources~\cite{zenon_information-theoretic_2019}, as the number of states become lower. Thus, it follows that contact interaction should be easier and therefore privileged.
It is also interesting that from a developmental viewpoint, we observe a progressive grasping behavior going from crude grasping in 7 month-old babies to precision grip in 3/4 years old children.

However, from a robotic viewpoint, it is rather paradoxal since joint actions are the most difficult motions to perform, and are avoided by researchers.
It is also unintuitive that the more the devices interact with each other, the easier the control should be as we diminish the number of possible states.

To remove this contradiction, we can say that the identification of those constraints are not obvious.
%
% Moreover, although our demonstration was for static conditions, we think it remains valid also for dynamical ones in which stable attractors may represent also situations of low entropy. %This case will be considered later.
%
Therefore, this simple example describes that mutualization of resources can achieve an information gain (or and entropy reduction) by reducing the number of free states. However, algorithms that effectively use mutual information are still scarce and difficult to implement in robotics for control and sensory fusion.

{\bf Motor precision, and learning}. Let's take another example of motor control, but this time, with a motor of high resolution $\bf R_X$ such that its angle $\bf \theta$ is within the interval range $\bf R_X=[0,\, 1]$, see Fig.~\ref{motor_resolution} a).
We assume that the motor can change its level of stiffness. Variable stiffness changes the relative force level in Newton theory, but in terms of  information, it induces also a change in precision, see Fig.~\ref{motor_resolution} a) and b).
%
% The variable stiffness force

For instance,
one very unreliable motor controller can be designed by dividing simply the motor space into two sub-spaces, $\bf R_{Y_0}=[0, 0.5]$ and $\bf R_{Y_1}=[0.5, 1]$, so that we can generate sequences of low precision, with motor codes of large uncertainties, similar to a discrete binary code; e.g., $y=[0\, 1\, 0\, 0\, 0\, 1\, 1\, 0\, 0]$.
This large imprecision is similar to an unreliable controller with loose tension and stiffness;  see Fig.~\ref{motor_resolution} b) on the left.

% motor hierarchy
The spatio-temporal motor grouping and their quantization permit to stabilize the motor synergies at various hierarchical level. For instance, the discrete motor pairing in Fig.~\ref{motor_resolution} a) constitutes a repertoire of the complete 4 possibilities in magenta $[0\, 0]\, [0\, 1]\, [1\, 0]\, [1\, 1]$. However, as the length of the motor chunks augments into triplets and quadruplets, the combinatorics increase respectively to $2^3=9$ and to $2^4=16$ possibilities in cyan and in green, but the number of observed motor synergies stabilize to motor repertoires of 5 actions only, which is lower than $9$ and $16$ and slightly higher than $4$.
Similar to language and the construction of words from letters, the created motor hierarchies help to reduce variability, to organize coherent movement over time, and to exploit them as patterns to diminish the redundancies.

% motor compositionality
To augment the level of precision, we can divide the subspaces $\bf R_{Y_0}$ and $R_{Y_1}$ into smaller subspaces;  see Fig.~\ref{motor_resolution} b) on the right. %two smaller subspaces $R_{00}$ and $R_{01}$ for $R_0$, and $R_{10}$ and $R_{11}$ for $R_1$.
%
%The type of sequences, still similar discrete binary code, is as follows: e.g., $y=[00,\, 11,\, 01,\, 00,\, 00,\, 10,\, 11,\, 00,\, 01]$.
%
%
Its dimension $\bf R_Y \in N$ can be very high, which demands to have very accurate sensors and motors of same precision level.
% The multi-resolution process can continue to the required level of precision.
%
Instead, the compositionality of low resolution codes $\bf Y$ can recreate time series $\bf X$ of much higher resolution so that $\bf R_Y \ll R_X$. In the motor domain, complex and nonlinear motor signals $\bf X$ can be represented then by a small number $\bf k$ of codes $\bf Y$. $\bf k$ relates then to the desired level of precision and stiffness.
The number $\bf k$ of recruited motor units $\bf Y$ becomes a parameter of the system's controllability.
%
%The body seen as a communication channel,
The number $\bf k$ of recruited sensorimotor codes $\bf Y$ controls then the robot's information flow and bottleneck.

%According to our proposal, the variable stiffness of motor changes the quantity of the information to process, and reversely,
% The number of reccruted units $k$ becomes then a control parameter perform a control to a desired precision level.
% length required to control them.
% the forthcoming information flow.
%
% Hence, similar with the channel's capacity in communication theory, it follows that the body is seen at the same time the interface and the bottleneck of information flow.

{\bf Fitts law and motor bandwidth}.  This bottleneck of the information flow in action and control is well examplified in the Fitts law~\cite{fitts_information_1954}. For instance, Fitts discovered that actions are inherently limited by a relation between speed, distance and precision so that rapid and distal actions become unprecise; see Fig.~\ref{motor_resolution} c). This law, which represents also an index of difficulty and a cognitive charge~\cite{zenon_information-theoretic_2019}, follows a logarithmic scale so that actions have necessarily a nominal speed for a certain precision level.
The Fitts law expresses therefore that actions have a particular bandwidth to perform accurate motions. %, for which the accuracy can be modulated to certain limits.
Similar with the compositional codes presented in Fig.~\ref{motor_resolution} b) for sensory precision, actions can be seen then as the composition of unreliable codes, and the number of recruited motor units $\bf k$ relates then to their cognitive charge based on the source coding equation.
In line with the motor synergies paradigm, few synergies can be necessary to reach a high precision level due to the exponential scale. %for which their combining increases their precision at the same time. %, relates directly to a cognitive cost.

% Thus, instead of controlling motor stiffness, these informational codes may control directly precision, which is a more interesting parameter to control.
%
{\bf Motor adaptivity and development}. During development, infants confront the challenge of resolving informational issues arising from the immaturity and unreliability of their sensors, the weakness and softness of their muscles, and the multitude of degrees of freedom they encounter. Despite these apparent constraints of the human body, infants successfully execute motion behaviors with remarkable efficiency, seemingly acquired effortlessly~\cite{lerner_motor_2015}.

Therefore, adapting one's own actions in terms of precision level (Information Theory) instead of motor stiffness (Newton Theory) can be a better parameter for controllability if the task has some uncertainty and variability.
For instance, there is no need to be very accurate in dynamic and noisy environments. In such situation, it is better to adapt one's own accuracy, which means $\bf R_Y$ or the motor code length or complexity, to the variance or uncertainty level of the environment $\bf R_X$, see Fig.~\ref{motor_resolution} d) left.
%
% Here, an adaptation task can be seen as the calibrating of one's own actions to the precision-level required for the environment.

Similarly, adaptation under uncertainty is what infants are experiencing the most during development, see Fig.~\ref{motor_resolution} d) right.
%
% During development, the learning and adaptation under uncertainty is what infants may experience the most, see Fig.~\ref{motor_resolution} d).
%
% The infant's body becomes then the channel to interface internal and external dynamics for efficient communication exchange.
%From an IT viewpoint,
%
% The interfacing between internal dynamics and external dynamics represents then the first learning phase for efficient communication exchange.
%
Therefore, the EM paradigm complies partly with others in cognitive and developmental sciences such as information-seeking, artificial curiosity, active sampling~\cite{gottlieb_information-seeking_2013, gottlieb_towards_2018}, active inference~\cite{friston_action_2010, pezzulo_hierarchical_2018,pezzulo_generating_2023}, and empowerment~\cite{klyubin_empowerment_2005}, proposing that infants are constantly and actively seeking novel information and maximizing their actions and their sampling to perceive, adapt and learn faster during development.

In contrast from them, however, the EM paradigm departs by predicting that incremental growth during development has to be exponential and not linear.
For instance, control dexterity develops rapidly in early life, with competence exponentially growing to encompass skillful abilities like reaching, grasping~\cite{corbetta_reach--grasp_2018, corbetta_perception_2021}, postural balance~\cite{adolph_learning_2008, adolph_development_2018, thurman_changes_2019}, and abstraction for solving embodied problems~\cite{spelke_core_2007,baillargeon_core_2012, allen_lifelong_2023}, and language acquisition~\cite{dupoux_cognitive_2018}.

Hence, from an IT viewpoint, we propose that the infant developmental phase characterized by the infant's apparent \emph{random} movements, what is called motor babbling~\cite{lerner_motor_2015, corbetta_perception_2021}, serve to explore at an incredible rate what the body is capable to do, toward maximizing the efficient information exchange between brain and environment; i.e., acquisition of body image~\cite{pugach_brain-inspired_2019, hoffmann_editorial_2020, morasso_body_2021}.
In this view, the infant's body becomes then the channel to interface internal and external dynamics for efficient communication exchange.

Thus, the 'interfacing' between internal dynamics and external dynamics, informational embodiment, may represent then the first learning phase to explore the 'richness' of its potentialities, which means acquisition of its complete combinatorics.
By doing so, informational embodiment can provide a perfect fit between the body's complexity to the one of its encoding; i.e., the information equivalence between body and brain, and the causal alignment of the sensorimotor circuits toward higher level cognition~\cite{kuniyoshi_humanoid_2004, hoffmann_body_2022, pitti_digital_2022, pitti_search_2022, kanazawa_open-ended_2023}.

% [Kuniyoshi, Hoffman].
%
% This view is different from developmental theories that emphasize the causal alignment of the sensorimotor circuits based on prediction.
% [Kuniyoshi, Hoffman].

% The observed random motions may serve to align causally the sensorimotor circuits and to diminish the complexity and errors between them..
%
% From an information viewpoint, any congruent information is also redundant. Contingent, synchronous information should be distinctive and privileged in comparison with lesser coordinated motions. This paradigm is slightly different with Hebbian learning and its slogan ``neurons that fire together wire together``.

% Let's have two motors controlling an arm or one finger, and positionned in opposition, as agonistic and antagonist muscles.
% %
% We assume that the two motors can change their level of stiffness and therefore their precision-level during control.
% %
% To control the arm, we can have very precise, high resolution motor angles such that they are comprise within the interval range $R$.
% %
% As we have seen, low-precision systems can emulate high precision ones. Accordingly, short and discrete motor codes may describe precise motor angles of higher resolution in same fashion.
%
% For instance, using two states, on-off, binary codes can serve to control high resolution motors following $\log R_{Motors} < k \log R_{binary} $. The codelength $k$ is the minimal length required to control them.

\section{Discussion}

{\bf Information maximization.} %While robotics use digital processing for sensors and motors encoding, IT and the concept of information maximization are not considered as design principles to comprehend and define the efficiency of information flow between the body morphology, sensory encoding and control.
In human communication systems, maximizing information corresponds to the design of specific codes efficient for particular channels, with respect to their physical limits in terms of information capacity, redundancy and noise. Based on that, the most robust codes are also those that reach the Shannon limit in terms of information transfer.

As an original idea, we consider the body as a communication channel so that the morphology (e.g., the redundancy of the limbs), the motors and the sensors, their placement, their number, and their resolution have an impact on the way information is retrieved, and quantified. Hence, the body has an incidence on information structure and therefore on the algorithms used for perception and control.

This idea is in line with the concept of morphological computation in the strict sense~\cite{paul_morphological_2006}: the body is processing information so that it is possible to provide efficient codes for sampling, and control and designing its body purposefully for it.

Interestingly, the digital treatment of information provided by Claude Shannon shows that when entropy is maximized, the information capacity in algorithms, communication channels and storage devices grows exponentially; e.g., the resolution of binary codes grows exponentially for each bit added.
%algorithms and  maximizing entropy very low resolution codes like binary codes (unreliable or unprecise) can gain in resolution by an exponential factor law for each bit added.

Following this, our definition of informational embodiment is the maximization of the bodily resources to its full exploitation of information capacity for sensing, learning and control with respect to its own physical characteristics and limitations.

This new formalism can be useful for the design of robots, and the understanding of human embodiment.
For instance, efficient coding mechanisms should operate at all robotic levels to withstand discrepancies, encompassing not only those related to control but also pertaining to material, morphology, sensors, and their interaction with the environment.

{\bf Informational embodiment. } IT utilizes entropy to gauge the level of information transmitted in a communication channel, irrespective of its specific properties. Building upon this concept, we suggest that entropy can be employed to quantify the information exchange within any type of robot's body, regardless of its characteristics. %, as depicted in Fig.~\ref{informational_embodiment}.

While this proposal may seem straightforward, it introduces a fresh perspective on certain concepts and theories from IT that are not immediately apparent but can be reinterpreted for robotics:

1. in his theory of communication, Claude Shannon formulated the source coding theorem to transmit a message along a transmission channel~\cite{shannon_mathematical_1948}. Importantly, this theorem emphasizes that the message to be transmitted and the transmission channel are interconnected unequally through their respective quantities of entropy, such as the resolution of the message and the channel capacity. This inequality describes the interdependence of these two variables—the message and the messenger—as part of the same equation. %; refer to Fig.~\ref{informational_embodiment} I.
Translated to robotics, this implies that the robot's body \emph{and} the algorithmic part must be considered in tandem.

2. the sole quantity exchanged between the two is entropy, the currency of information. To be more precise, it is the logarithm of entropy that is exchanged. %; consult Fig.~\ref{informational_embodiment} II.
This implies that the information exchanged and the relationships among interacting systems follow a power-law scale. For instance, with just a few interacting elements, combinatorics can rapidly grow exponentially. As explored further, this result has been extensively utilized in human-made modern communication systems to approach the theoretical limit set by Shannon (digital processing), whereas it is often overlooked or concealed in robotics. Notably, power laws are pervasive in nature for manipulating large numbers~\cite{strogatz_infinite_2020}. We propose that these laws of large numbers might be decisive for a robot's embodiment and efficiency, particularly in integrating and controlling numerous interacting systems—an aspect that remains challenging for current roboticists.

3. Shannon's equation elucidates that entropic elements interact multiplicatively and can approximate high-entropic objects very rapidly but also collapse just as swiftly. %; see Fig.~\ref{informational_embodiment} III.
This is derived from Boltzmann's equation in Thermodynamics. In sensing, control, and learning, weak and unreliable elements (e.g., sensors, actuators, neurons) are all limited in resolution. Nevertheless, when considered as a group, they can approximate more complex objects and signals, achieving higher resolution. Efficiency, in this context, is attained by enhancing entropy, signifying an information gain, while redundancy corresponds to an information loss. For robotics, this implies (1) that the interaction and coordination of (weak) elements can become combinatorial, enhancing efficiency if properly arranged, and (2) that the organization of (weak) elements among each other is a crucial design principle for rapid sensing, dexterous control, and learning/memory organization.

4. at the conceptual level, transcribing this theorem to robotics allows us to perceive the robot's body as the communication channel for the information sent and received. %; view Fig.~\ref{informational_embodiment} IV.
Therefore, this equation considers the characteristics of the robot's body as an intrinsic part of it—where structure and function are intertwined. The robot's embodiment, encompassing its morphology, the arrangement of its sensors and motors, and their resolution, can be quantitatively defined to address the challenges of sensing and control as problems of transmission or communication. This equation also expresses that the bodily capacity of a robot/agent is necessarily linked with the task it has to perform, and once again, the only  manipulated quantity is entropy.

% We believe that this interpretation of Shannon's source and channel coding equations is innovative and can significantly influence the way we design robots and controllers in the future. In a recent paper, we applied this theorem to artificial neural networks, demonstrating that we can reach Shannon's capacity limit in terms of information storage per neuron within the neural network~\cite{pitti_digital_2022}. We showed that even unreliable neurons can convey maximal information if properly designed, and that each added neuron to the neural population increases its storage capacity exponentially.

Following this, we may correlate the level of embodiment of a system (a robot or a biological agent) with its degree of information flow~\cite{lungarella_mapping_2006, lungarella_information_2007, zahedi_quantifying_2013, ghazi-zahedi_morphological_2019}. As one counter-intuitive consequence of it, informational embodiment predicts the long-term superiority of soft robots and reservoir computing in comparison to current robotics and AI architectures if properly designed for information processing. This is because they correspond to higher entropic systems with larger combinatorics.

{\bf Morphological Computation. }
% In the last decade, Rolf Pfeifer introduced the groundbreaking idea of incorporating computation into embodiment~\cite{pfeifer_embodied_2004, pfeifer_how_2006} and further proposed the concept of Morphological Computation~\cite{hauser_towards_2011, hauser_role_2012, pfeifer_revolution_2012, nakajima_information_2015, pfeifer_cognition_2014, iida_soft_2016, laschi_lessons_2016}.
%
Morphological computation has been defined as ”computation obtained through the interaction of physical forms”~\cite{paul_morphological_2006}.
Taken literaly, morphological computation makes a direct link between body, information, codes and computation~\cite{pfeifer_embodied_2004, pfeifer_how_2006}.
The concept has been further developped to include reservoir computing, chaotic behavior and soft robotics~\cite{pitti_creating_2010, hauser_towards_2011, hauser_role_2012, pfeifer_revolution_2012, nakajima_information_2015, pfeifer_cognition_2014, iida_soft_2016, laschi_lessons_2016}.

Morphological Computation posits that motion control in animals is not solely a model-driven computation in the brain; instead, it involves orchestration by components at multiple levels, distributed across passive and active features throughout the body and controller scales. The concept emphasizes the role played by the body, considering aspects such as passive dynamics, body morphology, and the strategic placement of actuators and sensors for optimal information acquisition and control. Thus, Morphological Computation advocates considering both the design of the robot body (hardware) and the controller (software). This approach presents new challenges for soft robotics and bio-inspired control, aiming to achieve the agility observed in animals.

However, Morphological Computation falls short in detailing the algorithms for information processing.
Recently, some data-driven AI approaches for model-free learning have been proposed~\cite{laschi_learning-based_2023}, although effective, they heavily rely on data and computational resources during learning and inference, posing efficiency concerns.
Besides, evolutionary AI algorithms and swarm computing approaches from Artificial Life have explored morphological design for robots, focusing on control and perception~\cite{bongard_resilient_2006, brown_universal_2010,corucci_material_2016, benureau_morphological_2022}.
In comparison, the framework of informational embodiment can provide a direct link between body structure and its computational algorithm, emphasizing randomness and discreteness to shape information structure with entropy and redundancy.

{\bf Physical Reservoir Computing. }
% {\it \bf [TODO Kohei]}
%
Reservoir Computing (RC), which is a class of models of  neural networks at the frontier of chaos, exploits random connections and nonlinear units.
These random networks have been suggested as models to entail the rich dynamics and behaviors in the body and in the brain~\cite{pitti_quantification_2005, pitti_exploration_2006, hauser_towards_2011, nakajima_information_2015, nakajima_reservoir_2021, laje_robust_2013, hinaut_real-time_2013, enel_reservoir_2016}.

In contrast to deep networks that use back-propagation, RC learns by tuning only the readout part that connects the middle layer to the output.
For instance, Allen Hart and colleagues provided a proof of embedding that reservoirs are universal approximators~\cite{hart_embedding_2020}.

% Typically,

% The learning of recurrent neural networks uses a method called backpropagation through time based on gradient descent, where the internal connection weights of the neural network are tuned. In contrast, Reservoir computing learns by tuning only the readout part that connects the middle layer to the output.

% In this method, the middle layer consists of a dynamical system with large degrees of freedom having huge number of non-linear elements randomly connected, called the reservoir.

% The connection weights within the reservoir and the connection weights linking the input to the reservoir are generally not trained and are set to fixed values (it should be noted that this is just a typical setting, and various variations have been devised in the field).

Using the high dimensionality of its internal memory, RC has the potential to compensate the low precision of its sensors and motors.
Coupled to a physical system, like a soft robot, its dynamics can be utilized as computational resources~\cite{nakajima_exploiting_2018}. This framework, called physical reservoir computing, is being applied in various fields such as neuromorphic devices and soft robotics\cite{pitti_creating_2010, nakajima_exploiting_2014, akashi_unpredictable_2019,inoue_soft_2019}.

Hence, the combination of physical and reservoir computing opens up the path to exploit natural physical dynamics as a computational resource, which means the dynamics of soft body or internal organ can also act, in a sense, similarly to the brain. It is about information processing via dynamics and its resource issue.

% Physical reservoir computing allows direct discussion of the relationship between physics and computation or information processing.
%
Therefore, we see a complementary view between informational embodiment and physical reservoir computing.
For instance, the control done by high combinatoric of reservoirs can be replaced directly by the highly nonlinear dynamics of a soft robot~\cite{akashi_embedding_2024}.

All in all, concepts such as entropy, resolution, quantization and redundancy can help to elaborate a physical and information theory for embodiment AI and robotics.

% We hypothesize that the strong nonlinearities found in these models reveal high entropic states that can approximate the one of very dense categorization and memory encoding capabilities.

% We also suggest that reservoir computing and spiking neural networks, both composed of random matrices, represent a class of neural networks of high entropy and therefore of very compressive codes, for large memory capacity. Hence, they must represent an important class of algorithms for autonomous robotics and neuromorphic AI.
% , which can be constituted of spiking neurons or chaotic oscillators,

% Although a physical robot evolved in a continuous space, low resolution codes reduce the number of states and effective dimension of the robot to be controlled.

% As the concept of morphological computation
% we can see the body as a computer, the number of states induce the combinatorics it can process.

% having a brain of same combinatorics level as the body,
% its number of states can be categorized.

% in a passive arm/tentacle
% Kohei check the number of state
% we check the memory capacity
% with random dynamics, it stores memory and you can
% the important thing is echostate condition
% environment is important to function as a computer
% each time you inject, you should replicate the same pattern.
%
% Allen Hart
% embedding and approximation theorms for echo state networks

\subsection*{acknowledgements}
Alexandre Pitti thanks the program ''Exploration Japon`` 2023 of the French Embassy for the travel grant.
This work was supported by the French National Agency for Research (PEPR O2R, project AS1, ANR-22-EXOD-0005).
Finally, the authors would like to thank Rolf Pfeifer and Max Lungarella for their cheerful and memorable discussions and insights.
%
% \subsection{Learning}
%
% \subsection{Recovery of missing items}
%
%
% \subsection{High-level planning - generation of paths between two states}

\bibliography{embodied23}
\bibliographystyle{sn-nature}

\end{document}